%% file: main.tex
\documentclass[usenames]{article}

\usepackage{hyperref}
\usepackage{url}

\usepackage{amsmath,graphicx,mlspconf}
\usepackage{algorithm}
\usepackage{algpseudocode}
\input{math_commands.tex}

\usepackage{tikz}
\usetikzlibrary{backgrounds,fit,arrows,external, positioning, calc, arrows.meta, shapes.geometric, shapes.misc}
\usepackage{etoolbox,siunitx}
\robustify\bfseries
\usepackage{booktabs}
\usepackage{pgfplotstable}
\usepackage{xcolor}
\usepackage{colortbl}
\usepackage{multirow}
\usepackage{todonotes}
\usepackage{cleveref}
\usepackage{enumitem}

\newcommand{\picp}[1]{\text{PICP}_{\SI{#1}{\percent}}}
\newcommand{\mpiw}[1]{\text{MPIW}_{\SI{#1}{\percent}}}

\title{Probabilistic modeling of lake surface water temperature using a Bayesian spatio-temporal graph convolutional neural network}

\name{%
    \hspace{-3em}Michael Stalder$^{\mathsection}$\sthanks{Email to corresponding author: michael.stalder@bfh.ch.}
    \enskip Firat Ozdemir$^{\ddagger}$%
    \enskip Artur Safin$^{\mathparagraph}$%
    \enskip Jonas Sukys$^{\mathparagraph}$\sthanks{Affiliation throughout the conducted work.}%
    \enskip Damien Bouffard$^{\mathparagraph}$%
    \enskip Fernando Perez-Cruz$^{\ddagger}$%
}
\address{%
    $^{\mathsection}$ Bern University of Applied Sciences, Department of Engineering and Information Technology, Switzerland\\
    $^{\ddagger}$ Swiss Data Science Center, Switzerland\\ 
    $^{\mathparagraph}$ EAWAG: Swiss Federal Institute for Aquatic Science and Technology, Switzerland \\
    }

\begin{document}
% \ninept

\maketitle

\begin{abstract}
Accurate lake temperature estimation is essential for numerous problems tackled in both hydrological and ecological domains.
Nowadays physical models are developed to estimate lake dynamics; however, computations needed for accurate estimation of lake surface temperature can get prohibitively expensive.
We propose to aggregate simulations of lake temperature at a certain depth together with a range of meteorological features to probabilistically estimate lake surface temperature.
Accordingly, we introduce a spatio-temporal neural network that combines Bayesian recurrent neural networks and Bayesian graph convolutional neural networks.
This work demonstrates that the proposed graphical model can deliver homogeneously good performance covering the whole lake surface despite having sparse training data available.
Quantitative results are compared with a state-of-the-art Bayesian deep learning method.
Code for the developed architectural layers, as well as demo scripts, are available\footnote{\url{https://renkulab.io/projects/das/bstnn}}.
%at:\\ .
\end{abstract}
\section{Introduction}
\label{sec:intro}

Being able to predict the water temperature at certain depths in lakes can provide crucial information for hydrological and ecological studies. 
These temperatures can be physically modeled and predicted using hydrodynamic-biological simulations \cite{Baracchini.}.
However, spatial discretization steps must be reduced to produce accurate predictions when getting closer to the lake surface. 
Fine discretization comes at a high computational cost. 
Applying artificial neural networks (ANNs) as a data-driven approach to describe lake surface water temperature could provide sufficient prediction accuracy at a fraction of the computational cost.

In recent years, numerous works have been proposed using recurrent neural networks (RNNs) on various kinds of sequence data, including time-series, which can capture temporal patterns of varying history. 
Further, the idea of Bayesian neural networks (BNNs)~\cite{Blundell.2015} has been applied to RNNs, yielding Bayesian recurrent neural networks (BRNNs)~\cite{Fortunato.10042017}, which allow for probabilistic modeling of sequence data. 
In addition, graph convolutional neural networks (GCNNs) have gained a lot of attention recently due to their ability to capture relations in unstructured data \cite{Wu.2021}.
The concept of Bayesian GCNNs has been introduced for classification tasks \cite{Zhang.2019b, Pal.08112019}, and used in the context of spatio-temporal modeling \cite{Zhao_2019_ICCV,Fu.15102020}.

This work introduces a Bayesian spatio-temporal model leveraging BRNNs and graph convolutions. 
The model is applied for probabilistic predictions of the lake surface water temperature of Lac L\'eman (Lake Geneva), Switzerland/France. 
The temporal model is first validated for a specific lake location (i.e., only temporal) and then combined with graph convolutions to predict values for the whole lake surface (i.e., spatio-temporal). 
Experimental results are compared with a current state-of-the-art method~\cite{kendall2017what}, the advantages of the proposed method and the developed code are discussed.

\section{Bayesian Neural Networks}
\label{sec:bnn}

Bayesian neural networks extend the artificial neural network architecture (ANN) for Bayesian inference. In other words, the ANN architecture is used to compute one sample of a predictive distribution with one forward pass of the ANN. Bayesian inference is then achieved by evaluating the ANN many times to obtain many predictive samples which describe the predictive distribution accurate enough.

\subsection{BNN weight parametrization}
For concise notation, the following is explained for a single neural network weight $w$. 
As proposed in \cite{Blundell.2015}, the variational posterior distribution of one weight follows a univariate Gaussian distribution:
\begin{equation}
	w \sim q(w)=\mathcal{N}(w\vert\mu, \sigma)
	\label{eq:bnn_weights}\ ,
\end{equation}
where $\mu$ and $\sigma$ are the learnable parameters. 
To allow for gradient-based optimization with back-propagation the so-called ``reparameterization trick'' is applied:
\begin{equation}
	w = \mu + \sigma\ast\epsilon,\quad \epsilon \sim \mathcal{N}(0,1)\ .
\end{equation}
Additionally, the standard deviation is parameterized by ${\sigma=\log(\exp(\rho)+1)}$, such that $\sigma$ is always a non-negative number \cite{Blundell.2015}. 
Thus, the final weight parameterization
is given by:
\begin{equation}
	w = \mu + \log(\exp(\rho)+1)\ast\epsilon,\quad \epsilon \sim \mathcal{N}(0,1)\ ,
	\label{eq:bnn_weights_repar}
\end{equation}
where $\mu$ and $\rho$ are learned parameters separately for each network parameter $w$.

\subsection{Priors}
Priors on the model parameters are defined as $p(w)$.
Each weight adds a loss based on the KL-Divergence between posterior distribution $q(w)$ and prior distribution $p(w)$ to the optimization target:
\begin{equation}
	l_{\text{KL}} = \alpha_{\text{KL}}\cdot\mathrm{D}_{\text{KL}}(q(w)\Vert p(w))\ ,
\end{equation}    
where $\alpha_{\text{KL}}$ is a hyper-parameter. 
The strength of the prior can be adjusted based on selected prior distribution, prior parameters, and hyper-parameter $\alpha_{\text{KL}}$.
Posterior sharpening as proposed in \cite{Fortunato.10042017} adds another loss term to the cost function:
\begin{equation}
	l_{\text{PS}} = D_{\text{KL}}\big(q(w^\prime\vert w,x)\Vert p(w^\prime\vert w)\big)\ ,
\end{equation}
where the posterior and prior distributions are given by:
\begin{align}
	q(w^\prime\vert w,x) &= \mathcal{N}(w^\prime\vert w-\eta\ast g_{w},\sigma_0^2) \label{eq:post_weights}\\
	p(w^\prime\vert w) &= \mathcal{N}(w^\prime\vert w, \sigma_0^2)\ ,
\end{align}
where $w$ is a BNN weight obtained from (\ref{eq:bnn_weights}), $\eta$ is a learnable parameter, $\sigma_0$ is a hyper-parameter, and $g_w$ is given by:
\begin{equation}
	g_{w} = -\nabla_w \log p(y\vert w,x)\ .
\end{equation}
For backpropagation, the reparameterization trick is applied to equation (\ref{eq:post_weights}). 
The parameters $\mu, \rho, \sigma$ are updated using gradients taken from a second evaluation of the loss function computed with a realization of weighs $w^\prime$ obtained from (\ref{eq:post_weights}).\\
A BNN with posterior sharpening requires two sampling steps per forward pass and has three learnable parameters per weight $w$ \cite{Fortunato.10042017}. 
Such a network is thus computationally heavier than the original Bayes by backpropagation approach \cite{Blundell.2015}.

\subsubsection{Calculation of KL-Divergence:}
The KL-divergence between continuous probability distributions is given by:
\begin{equation}
	D_\text{KL}(q(x)\Vert p(x)) = \int_{-\infty}^{\infty}q(x)\log\Big(\frac{q(x)}{p(x)}\Big) = \mathbb{E}\Big[\log\frac{q(x)}{p(x)}\Big]\ .
\end{equation}
For certain distributions (e.g., univariate Gaussian) above integral can be solved exactly. 
However, often no exact solution exists, e.g.\ when applying Gaussian mixture priors. 
In these cases, an estimate can be obtained by Monte-Carlo-approximation:
\begin{equation}
	D_{\text{KL}}^{\text{mc}} = \frac{1}{M}\sum\limits_{i=1}^M \log \frac{q(x_i)}{p(x_i)}\ ,
	\label{eq:kl_mc}
\end{equation}
where $M$ samples are taken from the posterior distribution $q$.

\subsection{Bayesian Recurrent Neural Networks} %temporal model
\label{sec:brnn}

In this work, we use the BNN extension for exploiting temporal data through recurrent neural networks as introduced in \cite{Fortunato.10042017}. 
The long short-term memory (LSTM)~\cite{Hochreiter1997LongSM} architecture is extended to support Bayesian inference. 
Namely, all weights and bias parameters of the LSTM cell are sampled from Gaussian distributions as shown in \cref{eq:bnn_weights,eq:post_weights}.
We add a fully connected BNN after the BRNN layers to produce the final lake surface water temperature predictions.
We call this model Bayesian temporal neural network (BTNN). 

\section{Graph Neural Networks}
\begin{figure*}[t]
    \centering
	\input{./graphics/model.tex}
	\caption{The spatio-temporal model consists of three sub models: 
	1) A temporal model built with LSTM layers, 
	2) a spatial model using graph convolutions and 
	3) a fully connected layer to contract the feature dimension to LSWT. 
	The colored sections highlight the input subset a subsequent layer is acting on.}
	\label{fig:model_overview}
\end{figure*}
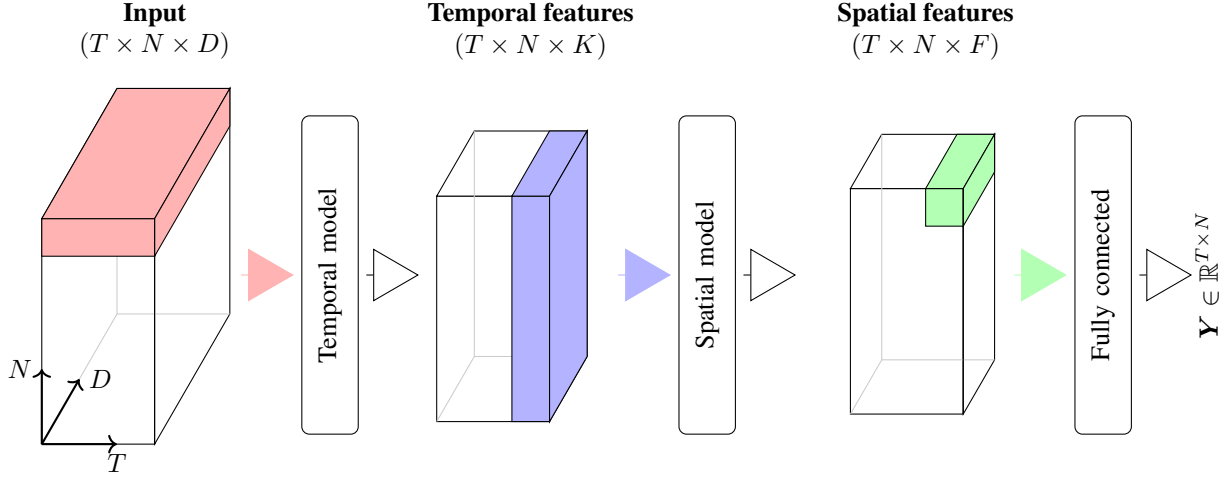
\label{sec:gcnn}

Convolutional neural networks are inherently limited to regular grid data such as regularly sampled time series, images, etc. 
Graph convolutional neural networks extend its applicable domain to data that reside on irregular grids. 
In graph convolutions, the grid neighborhood information is encoded in the adjacency matrix $\mA$ of graph $\gG$.
As shown in \cite{Kipf.09092016}, the following matrix operation approximates graph convolutions:
\begin{equation}
	\mZ_{i,t}^{(l)} = \tilde{\mD}^{-\frac{1}{2}}\tilde{\mA}\tilde{\mD}^{-\frac{1}{2}}\mH_{i,t}^{(l)} \bm\Theta^{(l)}
	\label{eq:graph_conv}
\end{equation}
where $\tilde{\mA}=\mA+\mI_N$, $\tilde{\mD}$ is the degree matrix of $\tilde{\mA}$, $\mH_{i,t}^{(l)}\in\R^{N\times C_{\mathrm{in}}}$ is the input for time step $t$ at layer $l$, ${\bm\Theta^{(l)}\in\R^{C_{\mathrm{in}}\times C_{\mathrm{out}}}}$ are the parameters to be learned, and $C_{\mathrm{in}}$ and $C_{\mathrm{out}}$ are the number of feature channels for input and output feature activations, respectively.
The layer propagation rule is completed by
\begin{equation}
	\mH_{i,t}^{(l+1)} = \sigma_\mathrm{act}(\mZ_{i,t}^{(l)})\ ,
\end{equation}
where $\sigma_\mathrm{act}$ is the activation function.

\subsection{Bayesian graph convolution}
\label{sec:bgcnn}
Formulating Bayesian graph convolution from \cref{eq:graph_conv} is straightforward by expressing $\bm\Theta$ with a Gaussian distribution:
\begin{equation}
	\bm\Theta \sim q(\bm\Theta)=\mathcal{N}(\bm\Theta\vert\bm\mu_\gG, \bm\sigma_\gG)\ .
	\label{eq:bayesian_gcnn}
\end{equation} 
To allow for back propagation, the distribution can be reparameterized using \cref{eq:bnn_weights_repar}.

\subsection{Graph construction for spatial data}
The easiest way to build the adjacency matrix for spatial data is to consider a node $j$ a neighbor of node $i$ if the Euclidean distance $d(i,j)$ is below a threshold. However, with such a graph construction, the influence of neighbors is independent of the Euclidean distance to a given center node. A more suitable way to define the graph structure for spatial data is to use a Gaussian diffusion kernel \cite{Henaff.17062015}:
\begin{equation}
	\tilde{a}_{ij}=\exp\big(-\dfrac{d(i,j)}{\sigma_\mathrm{dk}^2}\big)\ ,
\end{equation}
where $\sigma_\mathrm{dk}^2$ is a hyper-parameter.

\subsection{Bayesian spatio-temporal neural network}

In order to exploit spatial correlations within the lake, we extend the BTNN introduced in \cref{sec:brnn} with Bayesian GCNs (\cref{sec:bgcnn}). 
In particular, we decouple temporal evolution and spatial correlations, see \cref{fig:model_overview}. 
In our model the BTNN is shared among the spatial locations, which implies that the same sample of BTNN weights $\vw$ is used for all nodes, every time step, and all samples in one forward pass of the model, see \cref{alg:fp_lstm}. 
This is essential to ensure capturing spatial and temporal patterns.
The Bayesian GCN acts on the spatial dimension and its weights are thus shared for all time steps. 
It receives an input $\tH_i\in\R^{T\times N\times K}$ from the temporal model. However, the operation from equation (\ref{eq:graph_conv}) can be easily extended to higher dimensions:
\begin{equation}
	z_{itnf} = \sum_{k=1}^{K}\Big(\sum_{j=1}^{N} s_{nj}\cdot h_{itjk}\Big)\cdot \theta_{kf}\ , 
	\label{eq:graph_conv_2}
\end{equation}
where $\mS = \tilde{\mD}^{-\frac{1}{2}}\tilde{\mA}\tilde{\mD}^{-\frac{1}{2}}$, and $s$ and $h$ iterate indices of $\mS$ and $\tH$.
We call this model Bayesian spatio-temporal neural network (BSTNN). 

\begin{algorithm}[h]
\small
	\caption{Forward pass in spatio-temporal model.}
	\label{alg:fp_lstm}
	\begin{algorithmic}[1]
		\Procedure{Spatio-temporal model}{$\tX_i$}
		\State $\bm\epsilon \sim \mathcal{N}(0, I)$ 
		\State $\vw \gets \bm\mu + \log(\exp(\bm\rho+1))\circ\bm\epsilon$\Comment{Set weights}
		\For{$n={1,\dots,N}$}
		\State $\mX_{i,n}\gets\tX_i[:,n,:]$ \Comment{$\mX_{i,n}\in\R^{T\times D}$}
		\State $\tH_i[:,n,:] \gets \text{BRNN}(\vw,\mX_{i,n})$\Comment{$ \tH_i\in\R^{T\times N\times K}$}
		\EndFor	
		\State $\tG_i \gets \text{SpatialModel}(\tH_i)$\Comment{$\tG_i\in\R^{T\times N\times F}$}
		\State $\mY_i \gets \text{FullyConnected}(\tG_i)$\Comment{$\mY_i\in\R^{T\times N}$}
		\State \textbf{return} $\mY_i$
		\EndProcedure\end{algorithmic}
\end{algorithm}

\section{Data}
\label{sec:materials}

Data used for this work consists of meteorological data sets from several sources.
Below, we denote their nature (temporal:$^\tau$, spatio-temporal:$^\chi$) and content.\\
% listed below:\\
\textbf{Buchillon sensor data} ($^\tau$):
The station sensors record the lake surface water temperature (skin temperature) and the lake temperature \SI{1}{\meter} below the surface (bulk temperature). 
Air temperature, wind speed, pressure, relative humidity, wind direction, and net short wave radiation are recorded at \SI{4}{\meter}, \SI{7}{\meter} and/or \SI{10}{\meter} above the water surface. 
Our data set consists of hourly values from years 2017 and 2018. \\
\textbf{MeteoSwiss} ($^\tau$ \& $^\chi$):
This data is generated by meteorological simulations interpolated over the whole lake grid. 
It provides model estimates for air temperatures \SI{2}{\meter} above the lake surface, fractional cloud cover, short and long wave radiation, relative humidity, pressure, eastward and northward wind at \SI{10}{\meter} above the surface, and total precipitation. \\
\textbf{Meteolakes bulk temperature} ($^\chi$):
This data set includes bulk temperature simulations at different lake depths obtained from the Meteolakes system \cite{Baracchini.}\footnote{\url{http://meteolakes.ch/}}. 
For this work, the bulk temperatures at \SI{1}{\meter} depth are considered. \\
\textbf{Remote sensing} ($^\chi$):
This data set includes lake surface water temperature (LSWT) retrieved from advanced very high resolution radiometer (AVHRR)~\cite{Lieberherr2017performance} from satellite imagery. 
The data is very sparse, i.e., simultaneously valid observations hardly occur across the entire grid, and even each grid location rarely ($<5\%$) has a value at a given time instance.

\section{Experiments \& Results}
\label{sec:results}

% compBNN
For comparison, we implement the BNNs proposed in~\cite{kendall2017what}, % with Bidirectional~\cite{schuster1997bidirectional} LSTMs.
where one can quantify two kinds of uncertainties. %; model uncertainty due to limited data coverage and uncertainty that is due to implicit noise within available training data, respectively.
This is achieved through (i) Monte Carlo (MC) dropouts~\cite{gal2016dropout}, which is obtained through multiple forward passes with dropout layers enabled for inference, and (ii) optimizing the model for the negative log-likelihood of a normal distribution by means of a secondary model output which serves to estimate the predicted variance.
Precisely, the compared model is optimized for the loss function; ${\mathcal{L} = 0.5\exp(-s)||y-\hat{y}||^2+ 0.5s}$, where $\hat{y}$ and $s$ are the predicted lake skin temperature and predicted log variance (i.e.,\ $s=\log \hat{\sigma}^2$) for a given observation. 
For evaluations, we estimate predictive uncertainty of the model as
% ${\hat{\sigma}^2_{\mathrm{total}} = \mathrm{Var}(\{\hat{y}_e\}_{e=1,...,E}) + \mathbb{E}[\{\exp(s_e)\}_{e=1,...,E}]}$, 
\[{\hat{\sigma}^2_{\mathrm{total}} = \sum_e^E \hat{y}_e^2/E - (\sum_e^E\hat{y}_e/E)^2 + 
\sum_e^E \exp(s_e)^2/E}\]
for $E=11$ forward passes.
We refer to this compared method as compared Bayesian neural network (compBNN). 

\subsection{Experimental setup}
\label{sec:exp_setup}
We perform experiments for a single spatial point (temporal data set) and the entire lake (spatio-temporal data set). 
The time interval of the data sets do not fully overlap, but each cover approximately 2 years time.
For all experiments, one complete year is separated for testing, and the remaining year is split as weekly chunks into training and validation sets.

BTNN is trained with the objective function
% $\mathcal{L}_\mathrm{BTNN} = |\hat{y} - y|^2 + \lambda_\mathrm{KL}l_\mathrm{KL} + \lambda_\mathrm{PS} l_\mathrm{PS}$,
\begin{equation}
\label{eq:BTNN}
    \mathcal{L}_\mathrm{BTNN} = l_\mathrm{MSE} + l_\mathrm{KL} + l_\mathrm{PS}\ ,
\end{equation}
where $l_\mathrm{MSE}$ is the mean squared error between the estimated and observed target values. 
Referring to the parameters of BTNN and BGCNN as $\bf W_T$ and $\bf W_S$, we compare the following approaches for training BSTNN.\\
% For BSTNN training, we test and compare different approaches.\\ % One of the following lines to be picked based on available space in the row.
% We compare different approaches for training BSTNN.\\
% We consider the following for training BSTNN.\\
\textbf{BTNN independent (BTNN):} 
A BTNN is trained by treating grid nodes independently; i.e., one training sample holds values of only one grid location, which is randomly selected, and there are no BGCNN layers; optimizes $\bf W_T$ with $l_\mathrm{MSE}$, $l_\mathrm{KL}$ and $l_\mathrm{PS}$. \\
% In this setup, there are no BGCNN layers to capture spatial relations.\\
\textbf{Pre-trained (PT):}
From BTNN, the trained LSTM layers are extracted and reused in the complete BSTNN model shown in \cref{fig:model_overview}. 
During training, the LSTM weights are fixed such that only the graph convolution and dense layer weights are learned; optimizes $\bf W_S$ with $l_\mathrm{MSE}$ and $l_\mathrm{KL}$.\\
\textbf{Fine tuning (FT):}
This approach builds on a spatio-temporal model trained with the PT approach. 
Pre-trained LSTM layers and the graph are jointly optimized for 10 epochs; optimizes $\bf W_T$ and $\bf W_S$ with $l_\mathrm{MSE}$ and $l_\mathrm{KL}$.\\
% Adam optimizer with a learning rate of $\eta = \num{1e-5}$ is used.\\
\textbf{Joint training (JT):} 
All model weights are randomly initialized and optimized jointly for 40 epochs; optimizes $\bf W_T$ and $\bf W_S$ with $l_\mathrm{MSE}$ and $l_\mathrm{KL}$.
\\

\subsection{Metrics}
To compare our experiments we use the root mean squared error (RMSE) and the coefficient of determination ($R^2$). 
We evaluate these with the median value of the predictive distribution. 
For temporal experiments, weekly $R^2$ values are aggregated, then their median is reported as $\tilde{R}^2$.
For spatio-temporal experiments, we consider the full temporal sequence due to the sparsity of target observations. 
Scores are then averaged across spatial points to obtain scalar values and are denoted as $\bar{R}^2$ and $\overline{\text{RMSE}}$.
% Furthermore, $\tilde{R}^2$ stands for the median of weekly $R^2$ values, and $\bar{R}^2$ and $\overline{\text{RMSE}}$ stand for scores averaged among spatial points.
Additionally, we use metrics known as prediction interval coverage probability (PICP) and mean prediction interval width (MPIW) \cite{Pearce} to assess the coverage performance of the predictive distributions. 
We obtain the lower and upper bound $\hat{y}^L_i$ and $\hat{y}^H_i$ of the predictive distribution as the quantiles $\alpha$ of the distribution conforming to $P(\hat{y}^L_i \leq \hat{y_i} \leq \hat{y}^H_i) \geq 1-\alpha $. 
The corresponding metrics are denoted as $\text{PICP}_\alpha$ and $\text{MPIW}_\alpha$.  

\subsection{Implementation}
For the BSTNN we identified a two-layer LSTM with 16 and 32 units respectively, trained with $T=36$ and $P=8$, using a Gaussian prior with a standard deviation of 1 %$\sigma=1$
and a weight $\alpha_{\text{KL}}=0.001$ to be best performing.
We use this model as a basis for the BSTNN, where we add two Bayesian graph convolution layers with 64 units each\footnote{An example of this setup is provided among the demo files provided in our repository.}. 
We use $\sigma_\mathrm{dk}^2=1000$ and $\sigma_\mathrm{act}=\mathrm{ReLU}$.

\subsection{Results}
\label{sec:results_2}

\textbf{BTNN}: 
In \cref{table:best} we compare our BTNN with compBNN on the temporal data set. 
Best results are achieved with the BTNN using the full feature set (Buchillon + MeteoSwiss + Bulk temperature). 
On the reduced feature set (without Buchillon), the BTNN is outperformed on several metrics, however RMSE is almost the same. %, \cref{table:best}. 
This might be explained by the fact that the BTNN hyper-parameters are optimized for the full feature set.

\begin{table}[t]
	\centering
	\caption{Test set prediction performance comparison for the temporal data set. B~=~Buchillon, MS~=~MeteoSwiss, BT~=~bulk temperature.}
	\label{table:best}
	\vspace{10pt}
	\begin{tabular}{lccc}\toprule
        Model & BTNN & BTNN & compBNN\cite{kendall2017what} \\ \midrule
        Features & B+MS+BT & MS+BT & MS+BT \\ \midrule
        {$\tilde{R}^2$} & 0.81 & 0.57 & 0.66 \\
        {RMSE} & 0.74 & 0.91 & 0.92 \\
        $\picp{75}$ & 0.85 & 0.72 & 0.80 \\
        $\picp{90}$ & 0.93 & 0.86 & 0.91 \\
        $\mpiw{75}$ & 1.76 & 1.75 & 2.09 \\
        $\mpiw{90}$ & 2.49 & 2.46 & 2.99 \\ \bottomrule
	\end{tabular}
\end{table}
\begin{figure*}[]
    \begin{minipage}[b]{0.245\linewidth}
      \centering
      \centerline{\includegraphics[width=4.8cm]{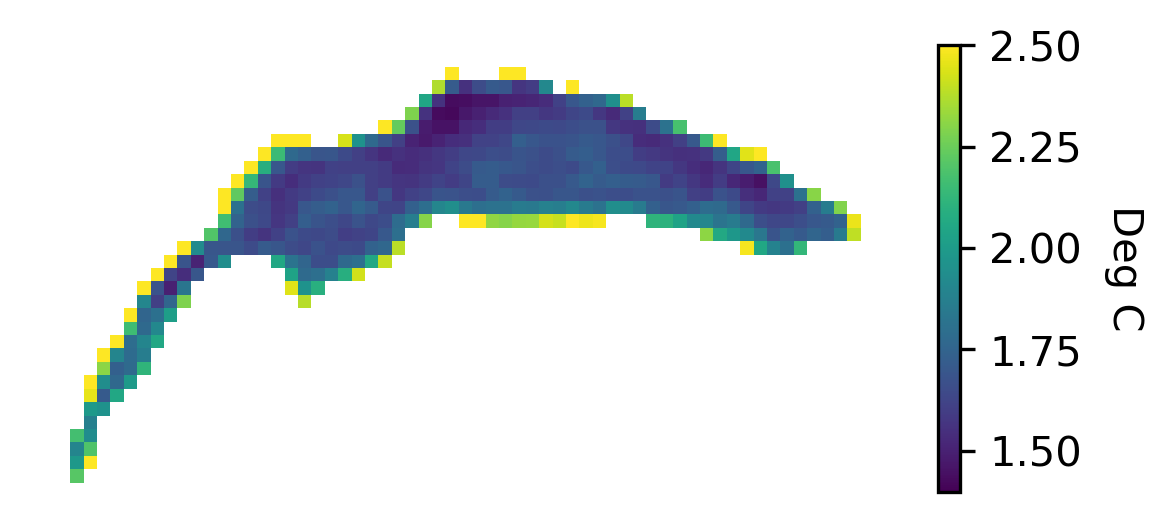}}
      \centerline{(a) BSTNN RMSE}\medskip
    \end{minipage}
    \hfill
    \begin{minipage}[b]{0.245\linewidth}
      \centering
      \centerline{\includegraphics[width=4.8cm]{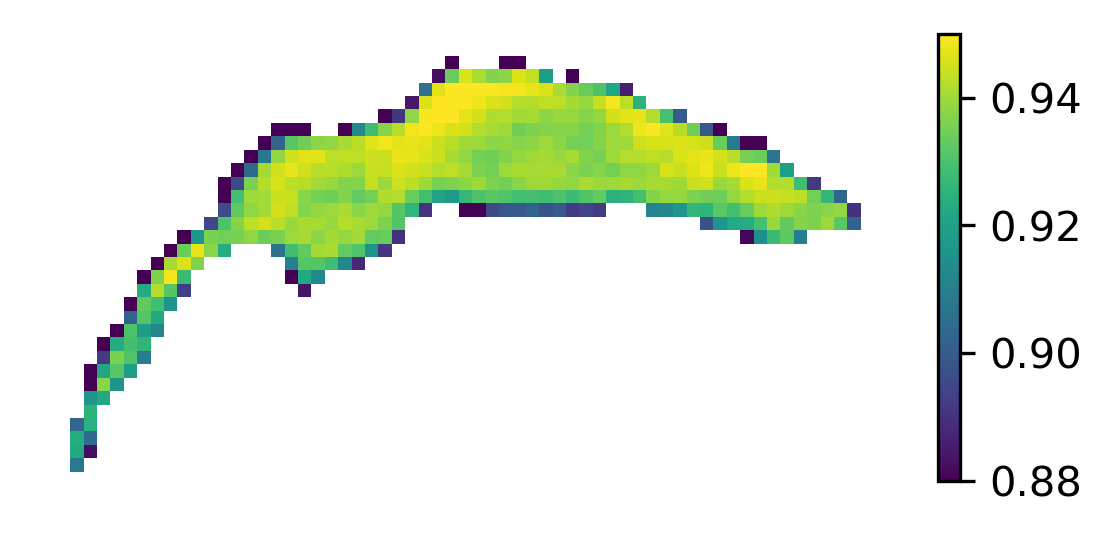}}
    %  \vspace{1.5cm}
      \centerline{(b) BSTNN $\bar{R}^2$}\medskip
    \end{minipage}
    \hfill
    \begin{minipage}[b]{0.245\linewidth}
      \centering
      \centerline{\includegraphics[width=4.8cm]{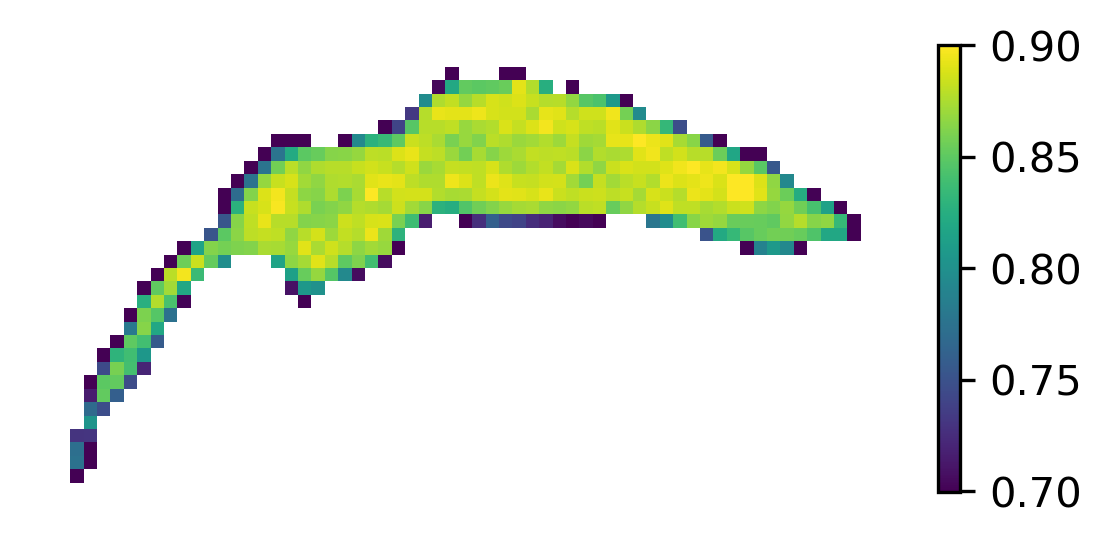}}
    %  \vspace{1.5cm}
      \centerline{(c) BSTNN $\picp{90}$}\medskip
    \end{minipage}
    \hfill
    \begin{minipage}[b]{0.245\linewidth}
      \centering
      \centerline{\includegraphics[width=4.8cm]{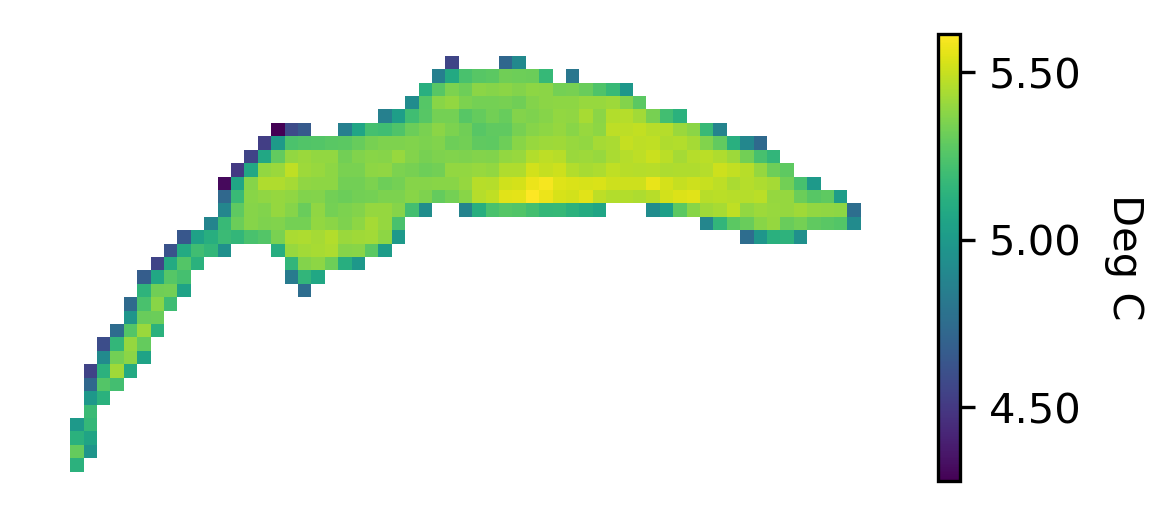}}
    %  \vspace{1.5cm}
      \centerline{(d) BSTNN $\mpiw{90}$}\medskip
    \end{minipage}
    %% row split
    \begin{minipage}[b]{0.245\linewidth}
      \centering
      \centerline{\includegraphics[width=4.8cm]{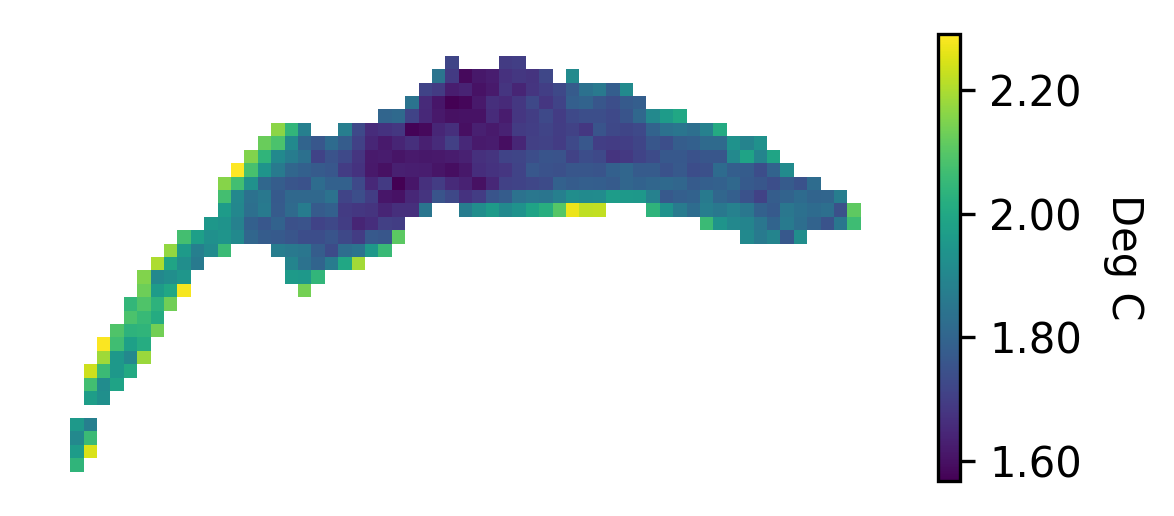}}
      \centerline{(a) compBNN RMSE}\medskip
    \end{minipage}
    \hfill
    \begin{minipage}[b]{0.245\linewidth}
      \centering
      \centerline{\includegraphics[width=4.8cm]{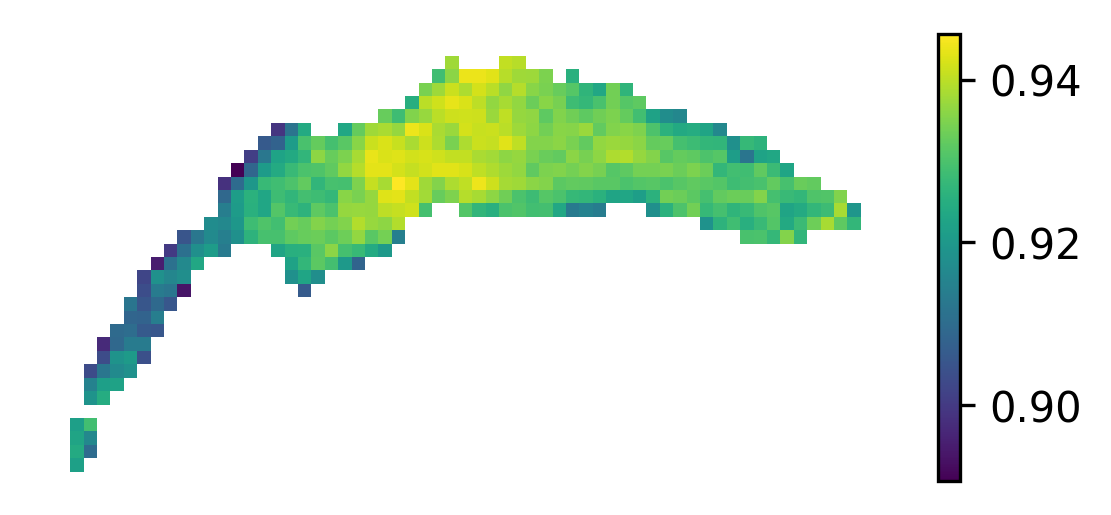}}
    %  \vspace{1.5cm}
      \centerline{(b) compBNN $\bar{R}^2$}\medskip
    \end{minipage}
    \hfill
    \begin{minipage}[b]{0.245\linewidth}
      \centering
      \centerline{\includegraphics[width=4.8cm]{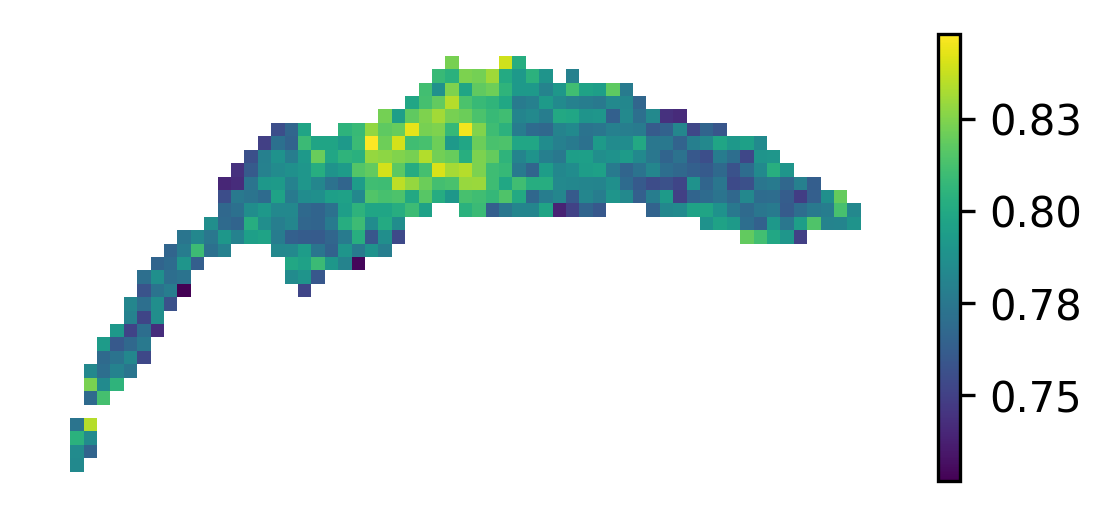}}
    %  \vspace{1.5cm}
      \centerline{(c) compBNN $\picp{90}$}\medskip
    \end{minipage}
    \hfill
    \begin{minipage}[b]{0.245\linewidth}
      \centering
      \centerline{\includegraphics[width=4.8cm]{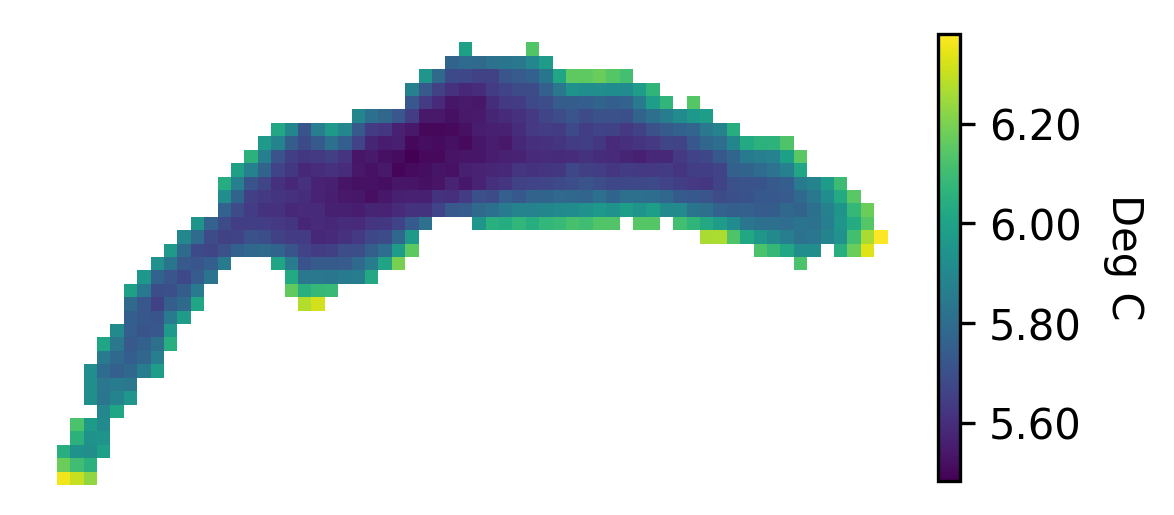}}
    %  \vspace{1.5cm}
      \centerline{(d) compBNN $\mpiw{90}$}\medskip
    \end{minipage}
    	\caption{Comparison of test set prediction performance for BSTNN (PT) and compBNN displayed over the lake grid.}
    	\label{fig:spatial_metrics}
\end{figure*}%

\noindent
\textbf{BSTNN}: 
\Cref{tab:training_approaches} compares the training approaches PT, FT, and JT. %, as described in \cref{sec:exp_setup}. 
Sequentially training the temporal and spatial models (PT) yields best results.  
Applying the same temporal model for all locations without a spatial model (BTNN) results in good accuracy metrics, however this approach results in an overconfident model with low PICP values. 
Also, fine tuning and joint training approaches do not achieve results as good as the PT approach. 
Accordingly, we use BSTNN PT for comparison with compBNN.

\begin{table}[b]
    \centering
    \caption{Validation set scores for different training procedures of the BSTNN for the spatio-temporal data set.}
    \label{tab:training_approaches}
    \vspace{10pt}
    \begin{tabular}{lllll} \toprule
     Metric\textbackslash Mode    & \textbf{BTNN} & \textbf{PT} & \textbf{FT} & \textbf{JT} \\ \midrule
        $\overline{\text{RMSE}}$ & 1.90 & 1.98 & 2.36 & 2.57 \\
        $\bar{R}^2$ & 0.84 & 0.83 & 0.76 & 0.71 \\
        $\picp{75}$ & 0.242 & 0.681 & 0.555 & 0.611 \\
        $\picp{90}$ & 0.329 & 0.816 & 0.675 & 0.793 \\
        $\mpiw{75}$ & 1.08 & 4.09 & 3.57 & 4.70 \\
        $\mpiw{90}$ & 1.51 & 5.78 & 5.11 & 6.53 \\ \bottomrule 
        
    \end{tabular}
    
\end{table}%

\begin{table}[t]
	\centering
	\caption{Test set prediction performance comparison for the spatio-temporal data set.}
	\label{table:test_scores}
	\vspace{10pt}
	\begin{tabular}{lcc}\toprule
    Model & BSTNN & compBNN\cite{kendall2017what} \\ \midrule
    {$\overline{\text{RMSE}}$} & 1.82 & 1.81 \\
    {$\bar{R}^2$} & 0.93 & 0.93 \\
    $\picp{75}$ & 0.72 & 0.62 \\
    $\picp{90}$ & 0.84 & 0.79 \\
    $\mpiw{75}$ & 3.70 & 4.04 \\
    $\mpiw{90}$ & 5.29 & 5.77 \\ \bottomrule
	\end{tabular}
\end{table} %

\Cref{table:test_scores} shows that BSTNN outperforms PICP of the reference method while having equal prediction accuracy. 
This is corroborated in \cref{fig:spatial_metrics}, which shows significantly more homogeneous and better prediction accuracy as well as PICP across the whole lake area for BSTNN, except for the lake shore.

\subsection{Discussion}
\label{sec:discussion}
In \cref{sec:results_2} we showed that our BTNN and BSTNN models perform equal or better than the reference method on the problem of lake water surface temperature prediction. 
In particular, the combination of Bayesian RNNs and graph convolutions show its strength in the presence of sparse data set.
The addition of graph convolutions improves the predictive performance over BTNN approach and the compBNN.
Although this is trivial to see for the scalar PICP and MPIW metrics in \cref{table:test_scores}, \cref{fig:spatial_metrics} depict the clear advantage of BSTNN for all metrics. 
This means that the predictive distributions captures the true value more often the other methods.
Our BSTNN generalizes better in the spatial domain compared to the reference method. 
All metrics are more stable across the lake, see \cref{fig:spatial_metrics}, with the exception of lake shore.
This can be attributed to (i) possibly less accurate bulk temperature simulations near shore and (ii) graph nodes at the shore having significantly less neighbors, and thus more weight is given to remaining neighbors, which might not correspond to the true physical phenomena. 
This could be addressed by treating shore nodes differently, e.g., by considering them as independent nodes without neighbors.
Looking at the histogram of valid LSWT measurements across the lake (figure not shown), we think that compBNN might be more dependent on having enough training samples available to produce good predictive distributions; since it achieves best results near the lake center, where most training samples are available.  

The implementation of Bayesian neural networks by formulating the variational posterior distribution of weights as Gaussian distributions used for BTNN and BSTNN has the advantage of offering more control over the learned distribution compared to the Monte-Carlo dropout method used in compBNN. 
Setting priors over parameters allow more control on the learned predictive distributions, and provide an additional tool set to encode prior knowledge when few samples are available.

\section{Conclusions}
\label{sec:conclusions}

In this work, we introduced a Bayesian spatio-temporal model combining Bayesian recurrent neural networks and graph convolutions, where quantitative analysis is conducted for the problem of lake surface temperature estimations. 
The temporal model (BTNN) is shown to predict the lake surface water temperature and its distribution accurately for a particular location on the lake. 
Compared to a state-of-the-art model, the spatio-temporal model (BSTNN) produced significantly more homogeneous predictions across the entire lake surface despite the sparse training data, showing clear signs of exploiting underlying spatial patterns.
\bibliographystyle{IEEEbib}
\bibliography{refs}

\end{document}

%% file: math_commands.tex
\usepackage{amsmath,amsfonts,bm}

\def\eqref#1{equation~\ref{#1}}
\def\1{\bm{1}}

\def\vw{{\bm{w}}}

\def\mA{{\bm{A}}}

\def\mD{{\bm{D}}}

\def\mH{{\bm{H}}}
\def\mI{{\bm{I}}}

\def\mS{{\bm{S}}}

\def\mX{{\bm{X}}}
\def\mY{{\bm{Y}}}
\def\mZ{{\bm{Z}}}

\DeclareMathAlphabet{\mathsfit}{\encodingdefault}{\sfdefault}{m}{sl}
\SetMathAlphabet{\mathsfit}{bold}{\encodingdefault}{\sfdefault}{bx}{n}
\newcommand{\tens}[1]{\bm{\mathsfit{#1}}}

\def\tG{{\tens{G}}}
\def\tH{{\tens{H}}}

\def\tX{{\tens{X}}}

\def\gG{{\mathcal{G}}}

\newcommand{\R}{\mathbb{R}}

%% file: graphics/model.tex
\newcommand{\drawbox}[8]{
    \pgfmathsetmacro \angle {30}
    \pgfmathsetmacro \xd {{#7*2/3*cos(\angle)}}
    \pgfmathsetmacro \yd {{#7*2/3*sin(\angle)}}
    \pgfmathsetmacro \width {{#5}}
    \pgfmathsetmacro \height {{#6}}
    \pgfmathsetmacro \x {{#1-\width+(#2-\width)*(\xd)}}
    \pgfmathsetmacro \y {{#3-\height+(#2-\height)*(\yd)}}

    \draw[fill=#4] (\x,\y) -- (\x+\width,\y) -- (\x+\width,\y+\height) -- (\x,\y+\height) -- cycle;
    \draw[fill=#4] (\x,\y+\height) -- (\x+\xd,\y+\height+\yd) -- (\x+\width+\xd,\y+\height+\yd) -- (\x+\width,\y+\height) -- cycle;
    \draw[fill=#4] (\x+\width,\y+\height) -- (\x+\width+\xd,\y+\height+\yd) -- (\x+\width+\xd,\y+\yd) -- (\x+\width,\y) -- cycle;
%    \draw[fill=#4] (\x,\y+\height) -- (\x+\xd,\y+\height+\yd) -- (\x+\xd,\y+\yd) -- (\x,\y) -- cycle;
    \draw[draw=#8] (\x,\y) -- (\x+\xd,\y+\yd) -- (\x+\width+\xd,\y+\yd);% -- (\x+\width,\y) -- cycle;
    \draw[draw=#8] (\x+\xd,\y+\yd) -- (\x+\xd,\y+\height+\yd);
	\draw (\x+\width,\y) -- (\x+\width,\y+\height);
}
\newcommand{\drawboxx}[7]{
	\pgfmathsetmacro \angle {60}
	\pgfmathsetmacro \xd {{#6*2/3*cos(\angle)}}
	\pgfmathsetmacro \yd {{#6*2/3*sin(\angle)}}
	\pgfmathsetmacro \width {{#4}}
	\pgfmathsetmacro \height {{#5}}
	\pgfmathsetmacro \x {{#1}}
	\pgfmathsetmacro \y {{#2}}

	\draw[fill=#3] (\x,\y) -- (\x+\width,\y) -- (\x+\width,\y+\height) -- (\x,\y+\height) -- cycle;
	\draw[fill=#3] (\x,\y+\height) -- (\x+\xd,\y+\height+\yd) -- (\x+\width+\xd,\y+\height+\yd) -- (\x+\width,\y+\height) -- cycle;
	\draw[fill=#3] (\x+\width,\y+\height) -- (\x+\width+\xd,\y+\height+\yd) -- (\x+\width+\xd,\y+\yd) -- (\x+\width,\y) -- cycle;
	%    \draw[fill=#4] (\x,\y+\height) -- (\x+\xd,\y+\height+\yd) -- (\x+\xd,\y+\yd) -- (\x,\y) -- cycle;
	\draw[draw=#7] (\x,\y) -- (\x+\xd,\y+\yd) -- (\x+\width+\xd,\y+\yd);% -- (\x+\width,\y) -- cycle;
	\draw[draw=#7] (\x+\xd,\y+\yd) -- (\x+\xd,\y+\height+\yd);
		\draw (\x+\width,\y) -- (\x+\width,\y+\height);
}

\tikzstyle{layer} = [draw,fill=white, minimum size=0.75cm, rounded corners=1mm, node distance=1.75cm, text width = 4cm, align=center]
\tikzstyle{label} = [fill = white, opacity=0, text opacity = 1, node distance=0.2cm, align=center]

\begin{tikzpicture}
	
	\coordinate (b1) at(0,0);
	\drawboxx{0}{0}{white}{1.5}{3}{3}{black!20}
%	\drawbox{1}{1}{-2}{red!10}{2}{0.5}{3}{none}
	\drawboxx{0}{2.5}{red!30}{1.5}{0.5}{3}{none}
	\node[label] (label1) [above = 5 cm of b1, xshift=1.5cm] {\textbf{Input}  \\ 
		$(T\times N\times D)$};

	\draw[->, thick] (0,0) -- (1,0) node[below] {$T$};
	\draw[->, thick] (0,0) -- (0,1) node[left] {$N$};
	\draw[->, thick] (0,0) -- (0.5,0.86603) node[right] {$D$};
%	\node (l1) at (0.75,-0.5) {time steps};
%	\node (l2) at (-3, 0) {grid nodes};
	\node[layer] (l1) [xshift=3.45cm, yshift=2.25cm, rotate=90, anchor=north] {Temporal model};
	\draw[red!30, arrows={-Triangle[angle=60:20pt]}] ($(l1.north)+(-0.8, 0)$) -- ($(l1.north)+(-0.1,0)$);
	\draw[arrows={-Triangle[angle=60:20pt, fill=white]}] ($(l1.south)+(0.1, 0)$) -- ($(l1.south)+(0.8,0)$);
	
	\drawboxx{5.25}{0.3}{white}{1.5}{3}{1.5}{black!20}
	\drawboxx{6.25}{0.3}{blue!30}{0.5}{3}{1.5}{none}
	\node[label] (label2) [right = 3.5 cm of label1.center, xshift=1.5cm, anchor=center] {\textbf{Temporal features}  \\ 
		$(T\times N\times K)$};
	
	\node[layer] (l2) [rotate=90, right=4.25 cm of l1.south, anchor=north] {Spatial model};
	\draw[blue!30, arrows={-Triangle[angle=60:20pt]}] ($(l2.north)+(-0.8, 0)$) -- ($(l2.north)+(-0.1,0)$);
	\draw[arrows={-Triangle[angle=60:20pt, fill=white]}] ($(l2.south)+(0.1, 0)$) -- ($(l2.south)+(0.8,0)$);
	
	\drawboxx{10.75}{0.4}{white}{1.5}{3}{1.25}{black!20}
    \drawboxx{11.75}{2.9}{green!30}{0.5}{0.5}{1.25}{none}
   	\node[label] (label3) [right = 3.75 cm of label2.center, xshift=1.5cm, anchor=center] {\textbf{Spatial features}  \\ 
    	$(T\times N\times F)$};
	
	\node[layer] (l3) [rotate=90, right=4.5cm of l2.south, anchor=north] {Fully connected};
		\draw[green!30, arrows={-Triangle[angle=60:20pt]}] ($(l3.north)+(-0.8, 0)$) -- ($(l3.north)+(-0.1,0)$);
	\node (outp) [right=1cm of l3.south, rotate=90, anchor=center] {$\mY\in\R^{T\times N}$};
	\draw[arrows={-Triangle[angle=60:20pt, fill=white]}] ($(l3.south)+(0.1, 0)$) -- ($(l3.south)+(0.8,0)$);
	
\end{tikzpicture}

%% file: main.bbl
\begin{thebibliography}{10}

\bibitem{Baracchini.}
Theo Baracchini,
\newblock {\em From observations to 3D forecasts: Data assimilation for high
  resolution lakes monitoring},
\newblock Phd, EPFL, Lausanne, 2019-05-24.

\bibitem{Blundell.2015}
Charles Blundell, Julien Cornebise, Koray Kavukcuoglu, and Daan Wierstra,
\newblock ``Weight uncertainty in neural networks,''
\newblock in {\em International Conference on Machine Learning}, Lille, France,
  2015, pp. 1613--1622.

\bibitem{Fortunato.10042017}
Meire Fortunato, Charles Blundell, and Oriol Vinyals,
\newblock ``Bayesian recurrent neural networks,''
\newblock {\em arXiv preprint arXiv:1704.02798}, 2017.

\bibitem{Wu.2021}
Zonghan Wu, Shirui Pan, Fengwen Chen, Guodong Long, Chengqi Zhang, and
  Philip~S. Yu,
\newblock ``A comprehensive survey on graph neural networks,''
\newblock {\em IEEE transactions on neural networks and learning systems}, vol.
  32, no. 1, pp. 4--24, 2021.

\bibitem{Zhang.2019b}
Yingxue Zhang, Soumyasundar Pal, Mark Coates, and Deniz Ustebay,
\newblock ``Bayesian graph convolutional neural networks for semi-supervised
  classification,''
\newblock {\em AAAI Conference on Artificial Intelligence}, vol. 33, pp.
  5829--5836, 2019.

\bibitem{Pal.08112019}
Soumyasundar Pal, Florence Regol, and Mark Coates,
\newblock ``Bayesian graph convolutional neural networks using node copying,''
\newblock {\em arXiv preprint arXiv:1911.04965}, 2019.

\bibitem{Zhao_2019_ICCV}
Rui Zhao, Kang Wang, Hui Su, and Qiang Ji,
\newblock ``Bayesian graph convolution lstm for skeleton based action
  recognition,''
\newblock in {\em International Conference on Computer Vision}, October 2019.

\bibitem{Fu.15102020}
Jun Fu, Wei Zhou, and Zhibo Chen,
\newblock ``Bayesian spatio-temporal graph convolutional network for traffic
  forecasting,''
\newblock {\em arXiv preprint arXiv:2010.07498}, 2020.

\bibitem{kendall2017what}
Alex Kendall and Yarin Gal,
\newblock ``What uncertainties do we need in bayesian deep learning for
  computer vision?,''
\newblock {\em arXiv preprint arXiv:1703.04977}, 2017.

\bibitem{Hochreiter1997LongSM}
Sepp Hochreiter and J{\"u}rgen Schmidhuber,
\newblock ``Long short-term memory,''
\newblock {\em Neural computation}, vol. 9 8, pp. 1735--80, 1997.

\bibitem{Kipf.09092016}
Thomas~N. Kipf and Max Welling,
\newblock ``Semi-supervised classification with graph convolutional networks,''
\newblock in {\em International Conference on Learning Representations}, April
  24-26, 2017.

\bibitem{Henaff.17062015}
Mikael Henaff, Joan Bruna, and Yann LeCun,
\newblock ``Deep convolutional networks on graph-structured data,''
\newblock {\em arXiv preprint arXiv:1506.05163}, 2015.

\bibitem{Lieberherr2017performance}
Gian Lieberherr, Michael Riffler, and Stefan Wunderle,
\newblock ``Performance assessment of tailored split-window coefficients for
  the retrieval of lake surface water temperature from {AVHRR} satellite
  data,''
\newblock {\em Remote Sens.}, vol. 9, no. 12, 2017.

\bibitem{gal2016dropout}
Yarin Gal and Zoubin Ghahramani,
\newblock ``Dropout as a bayesian approximation: Representing model uncertainty
  in deep learning,''
\newblock in {\em International Conference on Machine Learning}. PMLR, 2016,
  pp. 1050--1059.

\bibitem{Pearce}
Tim Pearce, Alexandra Brintrup, Mohamed Zaki, and Andy Neely,
\newblock ``High-quality prediction intervals for deep learning: A
  distribution-free, ensembled approach,''
\newblock in {\em International Conference on Machine Learning}. PMLR, 2018,
  pp. 4075--4084.

\end{thebibliography}
